\documentclass[10pt,twocolumn,letterpaper]{article}

\usepackage[pagenumbers]{cvpr} % To force page numbers, e.g. for an arXiv version
\usepackage{booktabs}
\usepackage{multirow}
\usepackage{makecell}  
\usepackage{threeparttable}
\usepackage{graphicx}  
\usepackage{amssymb} 
\usepackage{graphicx} 
\usepackage[utf8]{inputenc}
\definecolor{cvprblue}{rgb}{0.21,0.49,0.74}
\usepackage[pagebackref,breaklinks,colorlinks,allcolors=cvprblue]{hyperref}

\def\paperID{*****} % *** Enter the Paper ID here
\def\confName{CVPR}
\def\confYear{2025}

\title{FLUX-Text: A Simple and Advanced Diffusion Transformer Baseline for Scene Text Editing}

\author{
Rui Lan$^{*}$,\space\space
Yancheng Bai\footnotemark[1]\hspace{0.4em}\footnotemark[2] \space ,
Xu Duan,  
Mingxing Li,  \\
Dongyang Jin, \space
Ryan Xu, \space
Dong Nie, \space
Lei Sun, \space
and Xiangxiang Chu \\
{\normalsize Amap, Alibaba Group}\\
{\small \{lr264907, yancheng.byc, xuxu.dx, limingxing.lm, jindongyang.j\}@alibaba-inc.com}, \\
{\small ryansxu.00@gmail.com, \small dongnie@cs.unc.edu, \{ally.sl, chuxiangxiang.cxx\}@alibaba-inc.com} \\
{\tt  \small \url{https://amap-ml.github.io/FLUX-text/}}
}

\begin{document}

\maketitle

\footnotetext[1]{Equal contribution.}
\footnotetext[2]{Corresponding author.}
% \footnotetext[3]{Project Leader.}

\begin{abstract}
Scene text editing aims to modify or add texts on images while ensuring text fidelity and overall visual quality consistent with the background.
% Recent methods are built on latent diffusion models (LDM) and are primarily UNet-based, which have improved scene text editing results but still struggle with inaccurate or unrecognizable characters, especially for non-Latin ones (\eg, Chinese, Korean, Japanese) with complex glyph structures.
Recent methods are primarily built on UNet-based diffusion models, which have improved scene text editing results, but still struggle with complex glyph structures, especially for non-Latin ones (\eg, Chinese, Korean, Japanese).
To address these issues, we present \textbf{FLUX-Text}, a simple and advanced multilingual scene text editing DiT method. 
Specifically, our FLUX-Text enhances glyph understanding and generation through lightweight Visual and Text Embedding Modules, while preserving the original generative capability of FLUX.
We further propose a Regional Text Perceptual Loss tailored for text regions, along with a matching two-stage training strategy to better balance text editing and overall image quality.
Benefiting from the DiT-based architecture and lightweight feature injection modules, FLUX-Text can be trained with only $0.1$M training examples, a \textbf{97\%} reduction compared to $2.9$M required by popular methods.
Extensive experiments on multiple public datasets, including English and Chinese benchmarks, demonstrate that our method surpasses other methods in visual quality and text fidelity.
% All the code and model will be released.
All the code is available at \small\url{https://github.com/AMAP-ML/FluxText}.
\end{abstract}

\section{Introduction}
\label{sec:introduction}
Scene text editing aims to modify or add text in natural images while ensuring that the generated text remains accurate and is seamlessly integrated with the background~\cite{vtg2024}. 
This task has a wide range of applications, including advertisement design, poster updates, game scenes, and film post-production, and is of great significance to both professional designers and everyday users. 
However, it is highly challenging as it must handle multiple languages, fonts, sizes, and lines of text across complex and diverse visual contexts. 
The difficulty is particularly pronounced for non-Latin scripts such as Chinese and Japanese, where even subtle stroke omissions or glyph distortions can be easily perceived by humans.

\begin{figure}[!t]
\centering
\includegraphics[width=1.0\columnwidth]{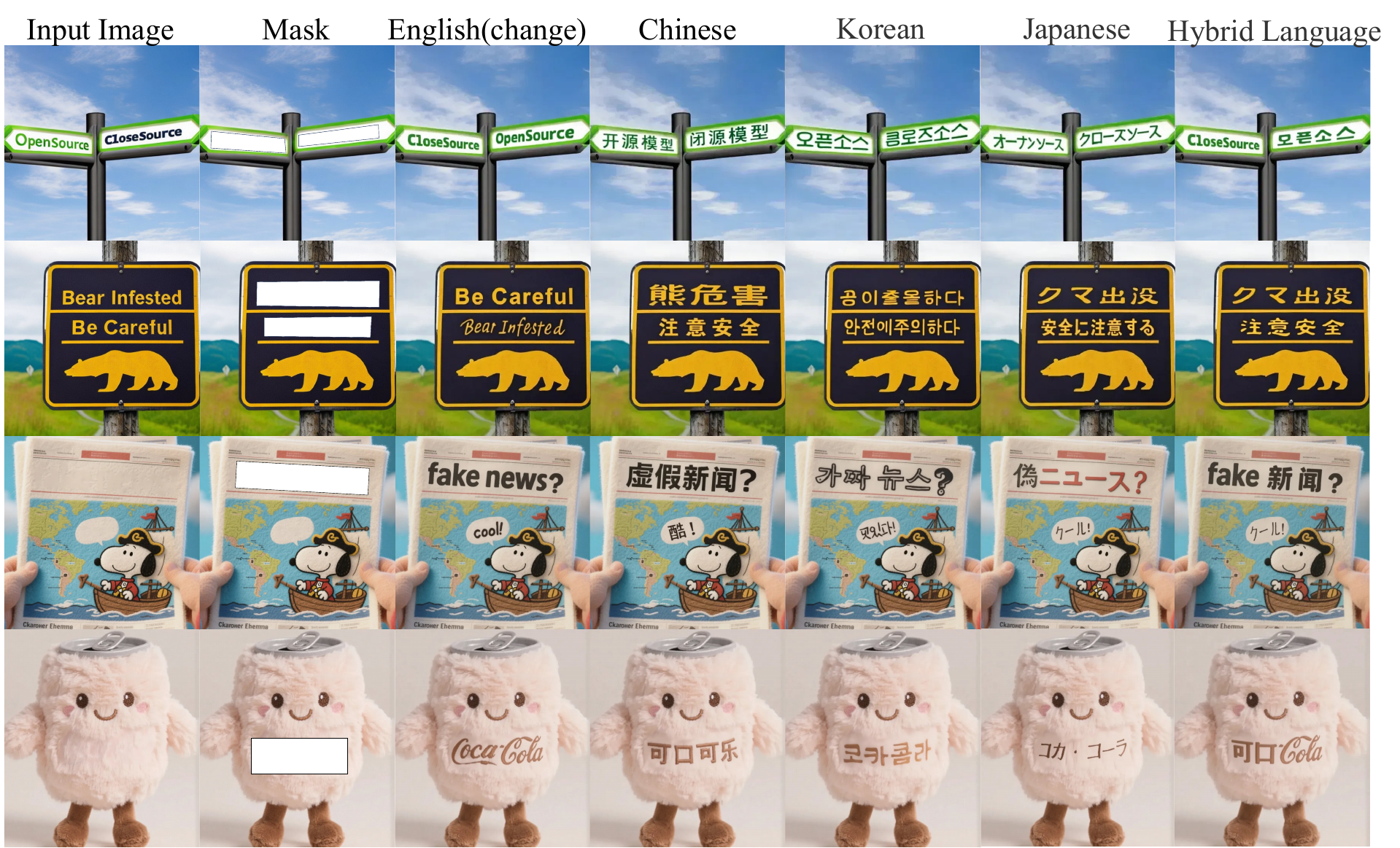}
\caption{
Scene text editing results of FLUX-Text under various conditions (e.g., English, Chinese, Korean, Japanese, and so on).
}
\label{fig:demo1}
% \vspace{-0.2cm}
\end{figure}

Recent large-scale text-to-image (T2I) diffusion models, such as FLUX~\cite{flux2024}, Stable Diffusion 3~\cite{SD32024}, and PlayGround~\cite{playground2_5_2024}, have demonstrated remarkable progress in open-domain image generation~\cite{playground2_5_2024}, producing realistic scenes with rich details. 
However, their performance in scene text editing reveals critical limitations. 
These models often fail to render text accurately, particularly in demanding scenarios that involve non-Latin languages with intricate glyphs (e.g., Chinese, with over 150K characters) or dense, multi-line layouts. The key reason is that these models lack explicit text-related priors and must ``learn from scratch'' how to render every character while also reasoning about font, color, layout, and other attributes, which frequently leads to errors and unnatural results, compromising both \textbf{visual quality} (seamless integration with the background) and \textbf{text fidelity} (structural correctness of glyphs).

To address the above issues, existing methods typically attempt to inject \textbf{text-editing cues} into diffusion models to enhance their text editing capability. 
These cues can be roughly divided into two categories:  
(1) \textbf{Visual embedding}, such as TextDiffuser-2 and AnyText2~\cite{anytext22024}, encode the visual appearance or positional layout of text and fuse it with background information into embedding vectors, which guide the model to generate the target text;  
(2) \textbf{Text embedding}, such as AnyText~\cite{anytext2023} and Seedream~\cite{seedream2025}, leverage OCR features or Glyph-ByT5-like semantic representations to constrain the generation process.  
Despite the effectiveness of these approaches, they still suffer from two major limitations. 
First, most of them are built upon ControlNet~\cite{controlnet2023} architectures, which introduce substantial additional learnable parameters and lead to low training efficiency. 
Second, their UNet-based backbones are inherently less capable of modeling complex visual contexts compared to the more advanced DiT~\cite{dit2023} architecture, resulting in inferior image quality and suboptimal text editing performance.

\begin{figure*}[!t]
\centering
\includegraphics[width=1.0\textwidth]{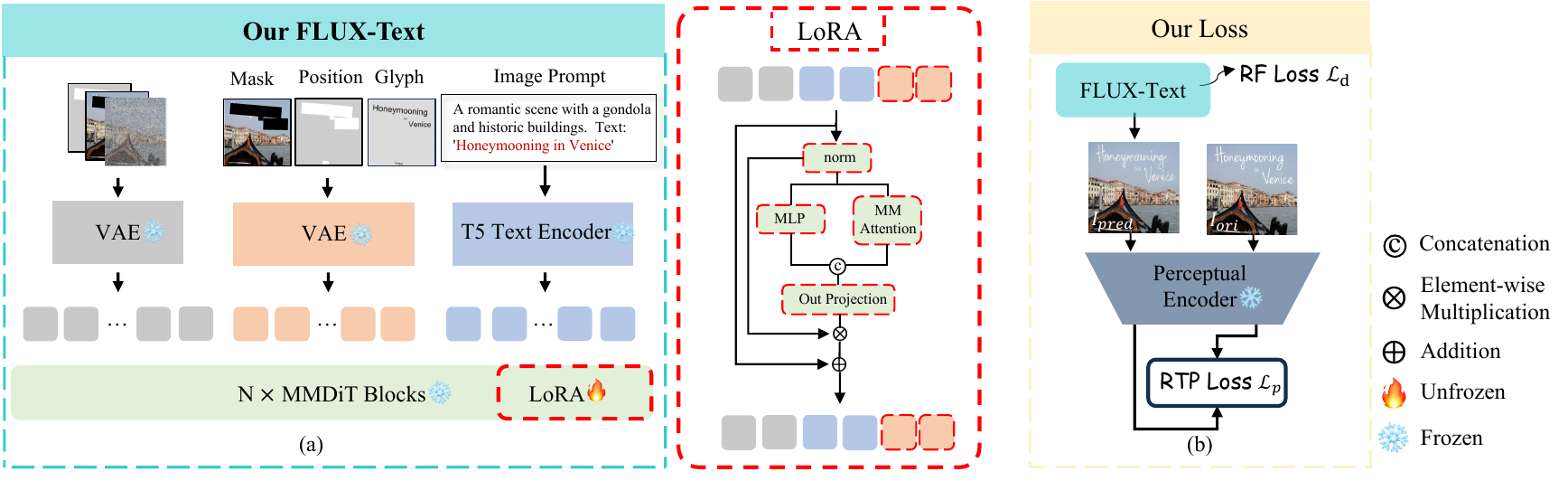}
\caption{
(a) The framework of our proposed \textbf{FLUX-Text}.
(b) Loss functions used to train FLUX-Text, including the RF loss and our proposed Regional Text Perceptual (RTP) loss.
}
\label{fig:network1}
% \vspace{-0.2cm}
\end{figure*}

In this paper, we propose a novel text editing model, named FLUX-Text, which achieves superior performance on text editing empowered by FLUX-Fill~\cite{flux2024}. 
Specifically, we extensively explore both the Visual and Text Embedding Modules and propose a new text-editing cues injection strategy that is particularly suitable for the DiT architecture, achieving an optimal balance between performance and efficiency.
Moreover, Regional Text Perceptual loss and a two-stage training strategy are proposed to guide the model focus on the text area, which not only ensures a harmonious integration of text and background but also significantly enhances the generation quality of the text regions.
% 增加创新点描述
Compared to other approaches~\cite{anytext2023,anytext22024,charGen2024,joyType2024}, our method achieves state-of-the-art (SoTA) performance on text editing tasks with only $100K$
training examples, reducing the training dataset by up to $97\%$. 
This efficiency mainly benefits from the DiT backbone and our efficient text-editing cues injection strategy.
The main innovations of our FLUX-Text framework are as follows:
\begin{itemize}
  \item To our knowledge, we are the first to apply a DiT-based framework to scene text editing. Building on extensive exploration of visual and semantic \textit{text-editing cues} injection, we propose \textbf{FLUX-Text}, a simple and effective method for robust text editing in complex scenes.
  % \item Regional text perceptual loss and two-stage training strategy are deployed and significantly boost the text editing performance.
  \item Regional Text Perceptual loss and the two-stage training strategy are applied to FLUX-Text, significantly improving the quality of text regions while maintaining high quality of other areas.
  \item Our method demonstrates significant improvements over previous approaches across diverse metrics, achieving SoTA results on the AnyText-benchmark and on MARIO-Eval-edit (an eval benchmark we adapted from MARIO-Eval specifically for scene text editing).
\end{itemize}

\section{Related Work}
\label{sec:related_work}

\subsection{Text-to-Image Generation}
\label{sec:t2i_generation}
Diffusion models~\cite{ddpm_2020} have become the leading paradigm for text-to-image (T2I) generation, offering strong fidelity and diversity. 
Latent diffusion models (LDMs)~\cite{LDM2022} improved efficiency by operating in compressed latent space, enabling large-scale systems such as GLIDE~\cite{glide2021}, Imagen~\cite{Imagen2022}, and DALL-E2~\cite{dall-E2-2022}. 
% Recent advances have shifted from UNet to Vision Transformers (DiT)~\cite{dit2023,SD32024,chenpixart,flux2024} and adopted stronger text encoders~\cite{radford2021learning,t5_2023} for better semantic alignment. 
Recent advances have also shifted the backbone architecture from conventional UNet designs to Vision Transformers (DiT)\cite{dit2023,SD32024,chenpixart,flux2024}, allowing for stronger global context modeling and more flexible adaptation to complex layouts. In parallel, the adoption of more expressive text encoders\cite{radford2021learning,t5_2023} has further improved semantic alignment between the textual prompt and generated content, addressing a key limitation of earlier systems.
FLUX~\cite{flux2024} builds on these trends with a transformer-based design and flow-matching objectives, providing a strong foundation for many downstream tasks such as controllable generation~\cite{song2023fashion,zeng2024cat,scalar,zhou2024realisdance,shi2024preference,chen2025finger,zhou2025realisdance} and image editing~\cite{anytext2023,ma2024,zhang2025context,zhang2025boow,zhang2025robust,chu2025usp,li2025ld}.

\subsection{Text Generation and Editing}
\label{sec:text_generation}
Generating and editing visual text in images remains challenging for diffusion-based models due to diverse fonts, complex backgrounds, and variable layouts.
% Recent methods have primarily focused on improving text generation by leveraging glyph conditions~\cite{glyphdraw2023,glyphcontrol2023}, segmentation masks~\cite{textdiffuser2023,anytext2023}, or character-aware text encoders~\cite{zhao2024udifftext,glyphdraw22025}. 
Recent methods have primarily focused on improving text generation by leveraging glyph conditions~\cite{glyphdraw2023,glyphcontrol2023}, segmentation masks~\cite{textdiffuser2023,anytext2023}, or character-aware text encoders~\cite{zhao2024udifftext,glyphdraw22025}, which provide additional visual priors and semantic information related to text editing.
Large language models (e.g., T5~\cite{t5_2023}) and transformer-based layout prediction~\cite{textdiffuser2023,textdiffuser22023} further enhanced spelling accuracy and complex text placement, while OCR supervision and pre-rendered glyphs~\cite{anytext2023,glyphdraw22025} improved character fidelity.
However, most approaches~\cite{anytext2023,anytext22024,charGen2024,textdiffuser2023,textdiffuser22023,brushtext_2024,glyphdraw2023,glyphdraw22025} remain UNet-based, where the locality of convolution limits their ability to model global context and adapt flexibly to layout changes. 
These methods often rely on heavy auxiliary networks (e.g., ControlNet~\cite{controlnet2023}) and large labeled datasets, which increases computational cost and complexity.
We propose FLUX-Text, a DiT-based method for scene text editing that better models global context and layout changes than UNets, while requiring far fewer parameters and training data without relying on heavy auxiliary networks.

\section{Methods}
\label{sec:methods}
In this section, we present FLUX-Text (shown in~\Cref{fig:network1}), a DiT-based method for scene text editing. We first introduce the overall diffusion pipeline in ~\Cref{sec:preliminary}. 
Then, we explain the Visual Embedding Module (\Cref{sec:visual_embedding_module}) and Text Embedding Module (\Cref{sec:text_embedding_module}) for injecting visual priors and semantic text information. 
Finally, we present our proposed loss design tailored for scene text editing in \Cref{sec:r_p_loss}.

\subsection{Preliminary} 
\label{sec:preliminary}
The text editing problem can be organized as follows: Given an image $X_i$, and a text prompt $y$, we have a set of text lines $T = \{(t_{1}, r_{1}), (t_{2}, r_{2}), \cdots, (t_{n}, r_{n})\}$. Here, $t_{j}$ represents the $j$-th text line to be edited within region $r_{j}$, and $n$ denotes the number of text lines. 
% To achieve high-fidelity and reliable editing results, our FLUX-Text utilizes the state-of-the-art T2I editing DiT model, FLUX-Fill~\cite{flux2024}.
We adopt the SoTA T2I editing DiT model FLUX-Fill~\cite{flux2024} as the baseline to ensure high-fidelity and reliable editing results.
Within the diffusion pipeline, both original image feature $z_0$ and mask image feature $z_m$ are derived by applying a VAE \cite{vaes2022} to the original input image $X_i$ and the mask image $X_m$ corresponding to the text regions. 
Meanwhile, text conditions $c_{te}$ are encoded by a T5 text encoder.
% To enhance text editing performance, the glyph-based conditions $c_{ve}$ and the text conditions $c_{te}$ are derived through the visual and text semantic embedding modules (implemented by a VAE and a T5 text encoder, respectively).
Subsequently, a noisy latent image feature $z_t$ is produced through a forward diffusion process, with $t$ denoting the time step. A DiT denoiser $\epsilon_{\theta}(\cdot)$ is utilized to estimate the noise added to the noisy latent image $z_t$ with the following objective:
\begin{equation}
\mathcal{L}_{d} = \mathbb{E}_{z_0, z_m, c_{te}, t \sim \mathcal{N}(0,1)} \left[ \| \epsilon - \epsilon_{\theta}(z_t, z_m, c_{te}, t) \|_{2}^{2} \right],
\end{equation}
% where $\mathcal{L}_{d}$ represents the rectified flow loss (RF loss)~\cite{SD32024}. 
% where $\mathcal{L}_{d}$ represents the rectified flow (RF) loss~\cite{SD32024}. 
where $\mathcal{L}_{d}$ means rectified flow (RF) loss~\cite{SD32024}.

% To enhance scene text editing, we need to effectively inject text-editing cues into the DiT model. To achieve this, we divide the text information into two complementary components: glyph-level visual priors (e.g., Canny edges and rendered glyph maps) and semantic text embeddings (e.g., OCR features or ByT5 representations). The former constrains text appearance, while the latter ensures semantic correctness, providing mutually reinforcing guidance. Thus, we explore multiple injection strategies in the visual and text embedding modules and determine the approach most suitable for the DiT architecture. 

\subsection{Visual Embedding Module}
\label{sec:visual_embedding_module}
Scene text editing aims to modify image glyphs with high visual fidelity, which is challenging for complex multi-stroke characters. 
This stems from the limited ability of existing methods to capture fine-grained visual cues, weakening the denoiser's guidance. 
To address this, Visual Embedding Module injects text-editing fine-grained visual cues to guide text editing.
In this module, we compare different image visual encoders and visual priors to determine the most suitable design for DiT-based scene text editing.

\begin{figure*}[!t]
\centering
\includegraphics[width=1.0\textwidth]{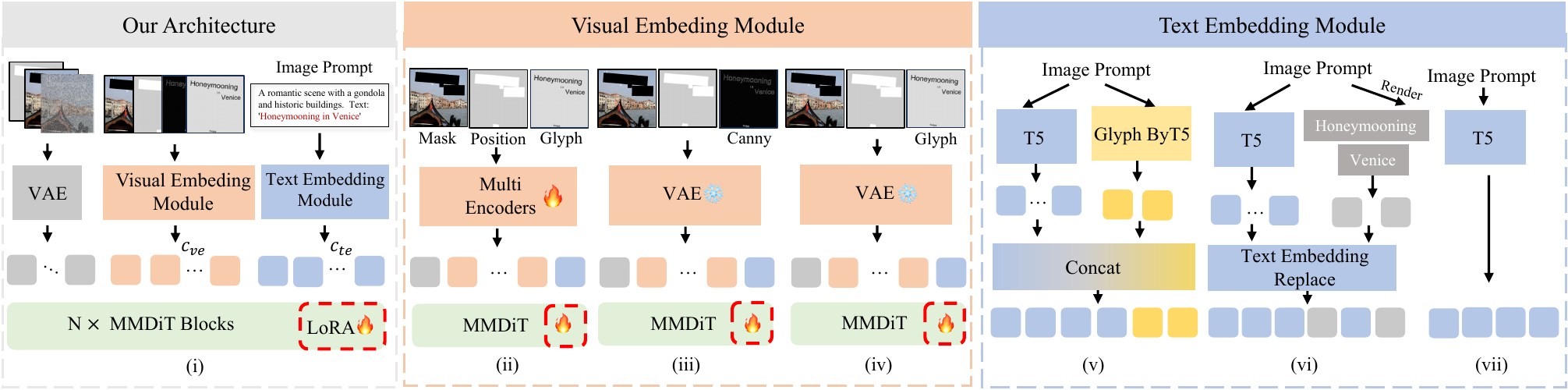}
\caption{
(i) Our Architecture.
(ii)$\thicksim$(iv) Different Visual Embedding Module.
(v)$\thicksim$(vii) Different Text Embedding Module.
}
\label{fig:network2}
% \vspace{-0.2cm}
\end{figure*}

\noindent \textbf{Different Visual Encoder.}
In this setting, three types of auxiliary conditions are utilized to produce latent feature map $c_{ve}$: Glyph $X_g$, Position $X_p$, Mask $X_m$. 
Specifically, Glyph $X_g$ is generated by editing texts $\{t_1, t_2, \cdots, t_n\}$ using a uniform font (\ie, `Arial Unicode') onto an image based on their locations $\{r_1, r_2, \cdots, r_n\}$. 
Position $X_p$ is generated by marking text positions $\{r_1, r_2, \cdots, r_n\}$ on an image.
Mask $X_m$ is simply obtained by masking out the text regions ${r_1, r_2, \cdots, r_n}$ from the original image.
We first need to encode all visual conditions and then inject them into the DiT framework. Consequently, the choice of visual encoder can significantly affect the feature quality and the final editing performance. Based on this, we explore different visual encoder designs to analyze their impact on editing performance.
% All visual conditions need to be encoded before being injected into the DiT framework, and the choice of visual encoder can significantly affect the feature quality and the final editing performance. Based on this, we explore different visual encoder designs to analyze their impact on editing performance.
\begin{itemize}
    \item \textbf{Multi-Encoders.} 
    In this design (\Cref{fig:network2}(ii)), each visual condition is encoded by an independent convolutional encoder ($\mathcal{G}$, $\mathcal{P}$, $\mathcal{M}$) trained separately. The encoded features of $X_g$, $X_p$, and $X_m$ are then fused through a convolutional layer $\mathcal{F}$ to produce the latent feature map $c_{ve}$:
    \begin{equation}
    c_{ve} = \mathcal{F}(\mathcal{G}(X_g) \mathbf{+} \mathcal{P}(X_p) \mathbf{+} \mathcal{M}(X_m)).
    \end{equation}
    While offering flexibility, this design introduces more parameters and higher complexity, which may lead to suboptimal performance.
    
    \item \textbf{VAE.}
    As shown in~\Cref{fig:network2}(iv), to minimize the number of learnable parameters, we directly use a frozen VAE encoder ($\mathcal{E}$) to extract features from the glyph $X_g$ and the mask $X_m$. 
    The position map $X_p$ is transformed by $\mathcal{R}$, a spatial rearrangement operator similar to the pixel-shuffle used in FLUX-Fill, to match the spatial size of $z_t$. 
    Finally, all visual conditions are mapped into the same latent space and concatenated to form $c_{ve}$:
    \begin{equation}
        c_{ve} = \operatorname{Concat}(\mathcal{E}(X_g), \mathcal{R}(X_p), \mathcal{E}(X_m)).
        \label{eq:glyph_vae}
    \end{equation}
    By aligning all visual conditions in a common latent space, this design injects well-aligned glyph conditions into the DiT denoiser without introducing additional parameters, leading to more stable and efficient training.
\end{itemize}

\noindent \textbf{Different Visual Priors.}
The choice of visual priors is critical for providing text-specific fine-grained cues to the DiT framework. 
In this setting, we investigate two types of visual priors for generating the glyph condition $X_g$: 
(i) edge-based priors such as Canny edges and 
(ii) rendered glyph maps using a uniform font (e.g., `Arial Unicode'). 
The edge-based priors are obtained by detecting the structural outlines of text regions, while the rendered glyph maps are generated by drawing the target texts $\{t_1, t_2, \cdots, t_n\}$ at their corresponding locations $\{r_1, r_2, \cdots, r_n\}$. 
We analyze how these priors affect text editing quality.
\begin{itemize}
    \item \textbf{Canny edges.} 
    In~\Cref{fig:network2}(iii), we embed edge-based glyph information by applying the Canny detector to $X_g$ and encoding the resulting edge map and the mask $X_{m}$ with a frozen VAE. The glyph condition $c_{ve}$ is then formed by concatenating the encoded edge features $X_{p}$, the reshaped position map $X_{p}$ and the mask $X_{m}$:
    \begin{equation}
    \label{eq:canny_injection}
    c_{ve} = \operatorname{Concat}(\mathcal{E}(\operatorname{Canny}(X_g)), \mathcal{R}(X_p), \mathcal{E}(X_m)).
    \end{equation}
    However, as noted in~\cite{charGen2024}, Canny often fails on small characters, which can degrade editing accuracy.
    \item \textbf{Glyph.}  
    We directly use the rendered glyph map $X_g$ in~\Cref{fig:network2}(iv), together with the position map $X_p$ and mask $X_m$, to form the glyph condition $c_{ve}$ as shown in eq.~\ref{eq:glyph_vae}. 
    Compared with edge-based priors, rendered glyph maps explicitly encode full character shapes and stroke details, providing richer structural information for accurate text editing.

\end{itemize}

% 加一段分析大体意思是Glyph+VAE

% \noindent \textbf{Glyph Injection.} 
% To make learned parameters as less as possible, we can directly use the VAE encoder to extract glyph information from $X_g$. We reshape $X_p$ to match the spatial size of $z_t$~\cite{flux2024}. Finally, we concatenate these two feature maps, resulting in $c_{ve}$, which can be represented as:
% \begin{equation}
% \label{eq:glyph_injection}
%   c_{ve} = \mathbf{Concat}(\mathbf{VAE}(X_g) , \mathbf{Reshape}(X_p)).
% \end{equation}
% In this injection, thanks to the strong encoding capability of VAE, sufficient glyph condition are injected into the DiT denoiser, and no extra parameters are introduced, thus enabling DiT to converge quickly with a smaller dataset. 

%% TODO Glyph injection 的名字是否合适，会让读者误解（conv 中有 glyph 的定义，下面没有定义）

\subsection{Text Embedding Module}
\label{sec:text_embedding_module}
Scene text editing also relies on accurate semantic representations of the target text to guide editing, as these embeddings help the DiT denoiser modify the text regions while preserving consistency with the background.
Previous works have combined different text encoders depending on the task objectives. 
For instance, CLIP embeddings were often paired with OCR features in UNet-based text editing frameworks~\cite{anytext2023}, 
while ByT5 was combined with T5 in DiT-based text-to-image generation models to strengthen semantic understanding~\cite{seedream2025}. 
Motivated by these designs, we explore three strategies (\Cref{fig:network2}(v)$\thicksim$(vii)) for injecting semantic text-editing cues from the caption $y$ into the DiT for scene text editing:

\noindent \textbf{T5 with OCR.}
% 添加一下为什么要T5+OCR
As shown in~\Cref{fig:network2}(vi), the processed caption $y'$, with editable text lines replaced by $S_*$, is tokenized and embedded via $\phi(\cdot)$, then fed into the pre-trained text encoder $\tau_{\theta}$ to obtain caption features.
The glyph lines are then rendered as images, passed through an OCR model $\gamma_{\theta}$ to extract features, and projected via an MLP $\xi(\cdot)$ to match the text encoder’s embedding dimension. 
Finally, these projected features replace the embeddings corresponding to $S_*$.
The textual representation $c_{te}$ integrates both text glyph and caption semantic elements and is expressed as follows:
\begin{equation}
\label{eq:ocr_step1}
c_{te} = 
\begin{cases}
\tau_{\theta}(\phi(y')) & t \neq S_* \\
\xi(\gamma_{\theta}(e_g)) & t = S_* ,
\end{cases}
\end{equation}
where $t$ denotes each token in the processed caption $y'$, where text lines to be edited are replaced by the placeholder $ S _*$. 
For $t \neq S_*$, we use the token embeddings from $\tau_{\theta}(\phi(y'))$. 
For $t = S _*$, the placeholder embedding is replaced by $\xi(\gamma_{\theta}(e_g))$, where $e_g$ is the rendered glyph image of the target text.

\begin{table*}[!t]
  \centering
  \caption{
  Evaluation results on AnyText-benchmark. 
  \( \ddagger \) represents the model trained on the full dataset.
  \textbf{Bold} indicates the best result and \underline{underline} indicates the second best.
  }
  \small
  \begin{tabular}{c|c|cccc|cccc}
    \toprule
    \multirow{2}{*}{Methods} & \multirow{2}{*}{Venue}         & \multicolumn{4}{c|}{English}                        & \multicolumn{4}{c}{Chinese}                        \\
                % \cmidrule{2-5} \cmidrule{6-9}
                & & Sen.ACC\(\uparrow\) & NED\(\uparrow\) & FID\(\downarrow\) & LPIPS\(\downarrow\) & Sen.ACC\(\uparrow\) & NED\(\uparrow\) & FID\(\downarrow\) & LPIPS\(\downarrow\) \\ 
    \midrule
    % FLUX-Fill~\cite{flux2024}   & 0.3093     & 0.4698    & 33.87        & 0.1582 & 0.0292     & 0.0625     & 29.93        & 0.1207               \\
    DiffSTE~\cite{diff-ste2023} &  Arxiv & 0.4523     & 0.7814    & 52.74        & 0.1816 & 0.0363     & 0.1226     & 57.49        & 0.1276               \\
    TextDiffuser~\cite{textdiffuser2023} & NeurIPS'23 & 0.5176     & 0.7618    & 29.76        & 0.1564 & 0.0559     & 0.1218     & 34.19        & 0.1252               \\
    DiffUTE~\cite{chen2023diffute} & NeurIPS'23 & 0.4054     & 0.7005    & 25.35        & 0.1640 & 0.2978     & 0.5744     & 29.08        & 0.1745               \\
    Anytext~\cite{anytext2023} & ICLR'24 & 0.6843     & 0.8588     & 21.59        & 0.1106 & 0.6476     & 0.8210     & 20.01        & 0.0943               \\
    TextCtrl~\cite{zeng2024textctrl} & NeurIPS'24  & 0.5853     & 0.8146    & 35.73        & 0.1978 & 0.3580     & 0.6084     & 49.79        & 0.2298               \\
    Anytext2~\cite{anytext22024} & Arxiv & 0.7915 & 0.9100 & 29.76 & 0.1734 & 0.7022 & 0.8420 & 26.52 & 0.1444 \\
    \midrule
    FLUX-Text\( \ddagger \) & \multirow{2}{*}{Ours} & \underline{0.8175} & \underline{0.9193} & \textbf{12.35} & \textbf{0.0674} & \textbf{0.7213} & \textbf{0.8555} & \textbf{12.41} & \textbf{0.0487} \\
    FLUX-Text &    & \textbf{0.8419} & \textbf{0.9400} & \underline{13.85} & \underline{0.0729} & \underline{0.7132} & \underline{0.8510} & \underline{13.68} & \underline{0.0541} \\
    \bottomrule
  \end{tabular}
  \label{tab:quantitative_comparison}
\end{table*}

\begin{table*}[!t]
\centering
    \caption{Evaluation results on MARIO-Eval-edit. 
    % Bold numbers indicate the best results. 
    % Notation is consistent with~\Cref{tab:quantitative_comparison}.
    \( \ddagger \) represents the model trained on the full dataset.
    \textbf{Bold} indicates the best result and \underline{underline} indicates the second best. 
    }
    \small
    \begin{tabular}{c|c|cccccc}
    \toprule
    Methods   & Venue    & FID\(\downarrow\)   & CLIPScore\(\uparrow\)  & OCR Accuracy\(\uparrow\) & OCR Precision\(\uparrow\) & OCR Recall\(\uparrow\) & OCR F1\(\uparrow\) \\ 
    \midrule
    % FLUX-Fill~\cite{flux2024} & 17.5204 & 0.2964   & 0.1714       & 0.3898        & 0.4129    & 0.4010 \\ 
    DiffSTE~\cite{diff-ste2023} &  Arxiv & 26.43 & 0.3134   & 0.1874       & 0.6511        & 0.7827    & 0.7109 \\ 
    TextCtrl~\cite{flux2024} & NeurIPS'24 & 15.12 & 0.3211   & 0.3722       & 0.7806        & 0.7412    & 0.7604 \\ 
    TextDiffuser~\cite{textdiffuser2023} & NeurIPS'23 & 13.49 & 0.3303   & 0.3609       & 0.7247        & 0.7203    & 0.7225 \\ 
    DiffUTE~\cite{chen2023diffute} & NeurIPS'23 & 8.55 & 0.3131   & 0.2684       & 0.6930        & 0.6879    & 0.6905 \\ 
    Anytext~\cite{anytext2023}    & ICLR'24 & 9.18  & \textbf{0.3343}   & 0.5015       & 0.7860        & 0.7847   & 0.7854 \\ 
    Anytext2~\cite{anytext22024}   & Arxiv & 15.16 & 0.3323   & 0.5705       & 0.8291        & 0.8212    & 0.8251 \\ \midrule
    FLUX-Text\( \ddagger \) & \multirow{2}{*}{Ours} & \underline{3.61}  & 0.3334   & \textbf{0.6667}       & \textbf{0.8798}        & \textbf{0.8800}    & \textbf{0.8799} \\ 
    FLUX-Text & & \textbf{3.11}  & \underline{0.3337}   & \underline{0.6510}       & \underline{0.8674}        & \underline{0.8755}    & \underline{0.8715} \\ 
    \bottomrule
    \end{tabular}
    \label{tab:mario}
\end{table*}

\noindent \textbf{T5 with ByT5.}
We encode editing texts with both the text encoder and a Glyph-ByT5 encoder $\delta_{\theta}$ to enhance holistic semantic information (as shown in~\Cref{fig:network2}(v)).
Then we employ an MLP layer $\eta$ to project the ByT5 embeddings into a space that aligns
with caption embeddings. Then, we concatenate both text glyph- and caption semantic features and obtain the final textual representation $c_{te}$, which is presented as follows:
\begin{equation}
\label{eq:GlyphByt5}
c_{te} = \operatorname{Concat}(\tau_{\theta}(\phi(y)), \eta(\delta_{\theta}(e_g))).
\end{equation}

\noindent \textbf{T5 Alone.}
In this setting (\Cref{fig:network2}(vii)), we directly use the pre-trained text encoder $\tau_{\theta}$ to process the caption $y$ without any additional glyph or auxiliary embeddings. 
The caption is tokenized and embedded via $\phi(\cdot)$, and the resulting token embeddings are passed through $\tau_{\theta}$ to obtain the final textual representation $c_{te}$:
\begin{equation}
\label{eq:T5_alone}
c_{te} = \tau_{\theta}(\phi(y)).
\end{equation}

\begin{figure*}[!t]
\centering
\includegraphics[width=2.0\columnwidth]{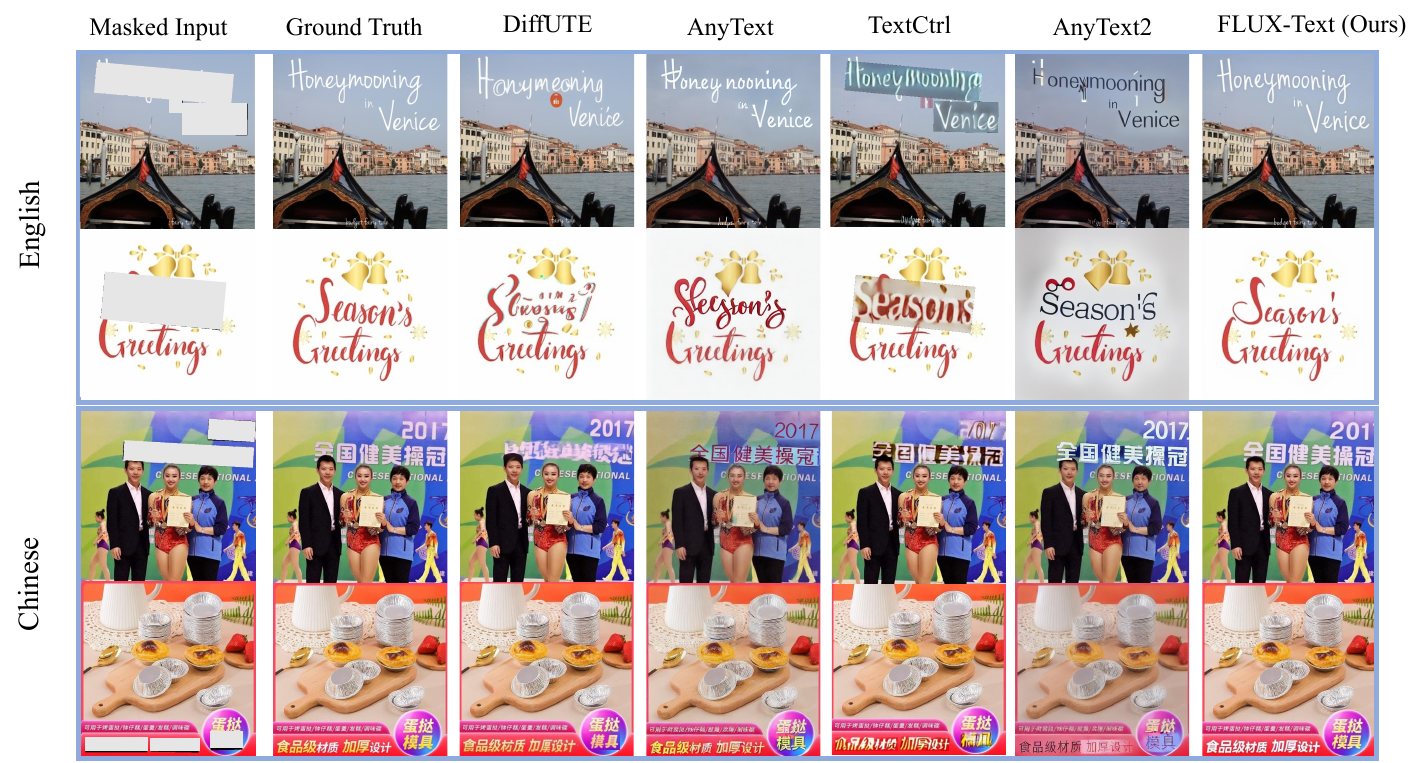}
\caption{Qualitative comparison of FLUX-Text and SoTA methods in Chinese and English scene text editing.}
\label{fig:result_vis}
%\vspace{-0.2cm}
\end{figure*}

\subsection{Regional Text Perceptual Loss}
\label{sec:r_p_loss}
Although the previous modules improved overall editing quality, the standard RF loss $\mathcal{L}_{d}$ still fails to capture fine-grained character details, making complex characters (e.g., Chinese glyphs) difficult to edit accurately due to their subtle strokes and structures~\cite{charGen2024,joyType2024}.
To address this, we improve the perceptual loss from~\cite{charGen2024} and introduce a Regional Text Perceptual (RTP) loss that focuses exclusively on text regions by applying a position mask $X_p$ during loss computation, thus providing stronger supervision for character-level details:

\begin{equation}
\label{eq:per_loss1}
    \mathcal{L}_{p} = \sum_{k} \frac{1}{\sum_{h,w}X_{p,{hw}}} \sum_{h,w} \| (f_{hw}^k  - \hat{f}_{hw}^k) \cdot X_{p,{hw}} \|_2^2,
\end{equation}
\begin{equation}
\label{eq:per_loss2}
    \hat{f}_{hw}^k = \text{TEncoder}(X_{i}), \quad \hat{f}_{hw}^k = \text{TEncoder}(X_{pred}),
\end{equation}
where $f_{hw}^k$ and $\hat{f}_{hw}^k$ are the $k$-th multi-scale feature maps extracted by $\text{TEncoder}$ from the predicted and ground-truth images, respectively.

The overall training loss is defined as:
\begin{equation}
\label{eq:final_loss}
    \mathcal{L} = \mathcal{L}_{d} + \lambda \cdot \mathcal{L}_{p},
\end{equation}
where $\lambda$ is a hyperparameter that that balances the RF loss and RTP loss.
Incrementing $\lambda$ directs the model to place greater emphasis on training the text area.

Naively setting $\lambda$ too high at the early training stage can cause convergence issues.
To address this, we adopt a \textbf{two-stage training strategy}: $\lambda$ is first set to a small value in the initial stage to allow the model to learn a stable foundation. Once convergence is achieved, $\lambda$ is increased in the second stage to place greater emphasis on text regions.
This strategy improves the integration of text and background and significantly enhances the quality of the generated text regions.

\subsection{Summary}
Based on the above analysis about~\Cref{fig:network2} and experimental validation in~\Cref{sec:comparison}, we finalize \textbf{FLUX-Text} (in~\Cref{fig:network1}) with the VAE encoder, rendered glyph priors from the Visual Embedding Module, the single-T5 text encoder from the Text Embedding Module, and the proposed Regional Text Perceptual Loss with a two-stage training strategy.
This design combines the stable, parameter-efficient feature alignment of the Visual Embedding Module with the strong semantic representations of the Text Embedding Module, achieving the best trade-off between accuracy, stability, and efficiency in DiT-based scene text editing.

\section{Experiment}
\label{sec:experiment}

\subsection{Implementation Details}
\label{sec:implementation_details}
Our proposed model is trained on a high-performance computing setup comprising $16$ H20 GPUs. 
The training procedure is extensive, spanning $20$K iterations over a period of approximately $2.5$ days. The training steps in stage 1 and stage 2 are $15$K and $5$K, respectively.
To optimize the training process, we employed the Prodigy optimizer \cite{prodigy2024},
integrating features such as safeguard warmup and bias correction to enhance stability and convergence. 
The weight decay parameter is set to $0.01$ to mitigate overfitting. 
The batch size is set as $256$ and training images are processed at a resolution of $512\times512$. This configuration balances the computational load and model accuracy effectively.

\begin{figure*}[!t]
\centering
\includegraphics[width=2.0\columnwidth]{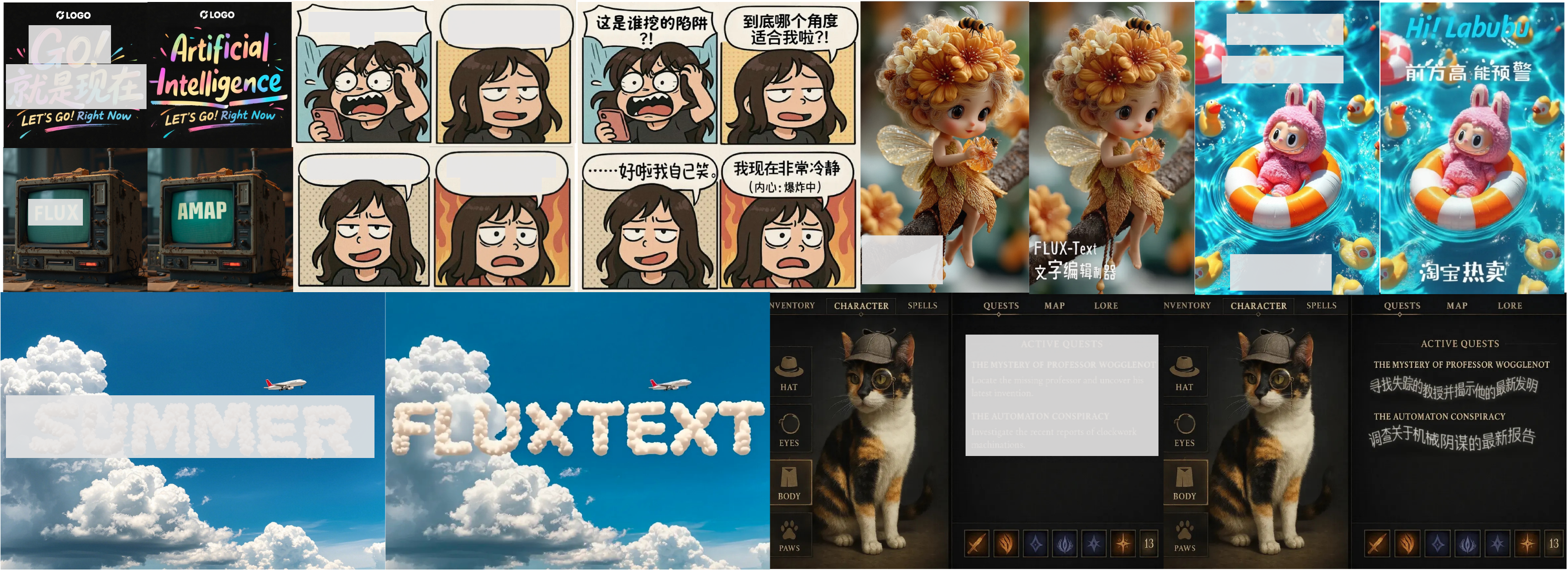}
\caption{
Visual generalization of FLUX-Text on web-crawled images (Left: masked input, Right: our result).
}
\label{fig:vis_zero}
\end{figure*}

\subsection{Dataset}
\label{sec:training_dataset}
\noindent \textbf{Training Dataset.}
% We utilize the AnyWord-3M dataset~\cite{anytext2023}, a comprehensive multilingual collection of publicly sourced images. 
% This dataset integrates images from several notable sources including Wukong~\cite{wukong2022}, LAION~\cite{laion2021}, and various datasets specifically tailored for OCR recognition tasks. 
% The images within this dataset depict an array of text-rich scenes, such as urban landscapes, book covers, advertisements, posters, and movie frames. 
% The dataset comprises approximately $3$ million images, with a linguistic 
% distribution of around $1.6$ million images featuring Chinese text, $1.39$ million containing English text, and an additional $10K$ images showcasing text in other languages. %, such as Japanese, Korean, Arabic, Bangla, and Hindi. 
% To train FLUX-Text, we build a small dataset of $100K$ images extracted from AnyWord-3M, comprising $50K$ Chinese images and $50K$ English images, respectively.
We utilize the AnyWord-3M dataset~\cite{anytext2023}, a large multilingual collection of $3$M publicly sourced, text-rich images from sources such as Wukong~\cite{wukong2022}, LAION~\cite{laion2021}, and several OCR-focused datasets.
It covers diverse scenes (e.g., urban landscapes, book covers, advertisements, and movie frames) with $1.6$M Chinese, $1.39$M English, and $10$K other-language images.
To train FLUX-Text, we deliberately construct a small $100$K image subset of AnyWord-3M ($50$K Chinese and $50$K English), demonstrating that our method can achieve strong performance even with significantly less training data.

\noindent \textbf{Testing Dataset.}
(i) The AnyText-benchmark~\cite{anytext2023} contains 1,000 images from Wukong~\cite{wukong2022} (Chinese) and LAION~\cite{laion2021} (English) for assessing text generation accuracy and quality.
(ii) MARIO-Eval-edit includes 4K image-text pairs from a subset of the MARIO-10M test set. 
It is adapted from MARIO-Eval~\cite{textdiffuser2023} to evaluate our scene text editing task, for which we extract the corresponding text masks and will release them together.

\subsection{Evaluation Metrics}
\label{sec:testing_dataset}
(i) For AnyText-benchmark, we adopt four metrics covering both text fidelity and visual quality.
Sentence Accuracy (Sen. Acc.) measures OCR correctness by cropping and recognizing each generated text line, while Normalized Edit Distance (NED) provides a more flexible string-level similarity. 
For generation quality, we use Fréchet Inception Distance (FID)~\cite{fid2020} and LPIPS~\cite{lpips2018} to assess distribution-level and perceptual similarity, ensuring style consistency between edited and unedited regions.
(ii) For MARIO-Eval-edit, we follow the evaluation metrics of TextDiffuser~\cite{textdiffuser2023} for all editing models.

\subsection{Comparison Results}
\label{sec:comparison}
\noindent \textbf{Quantitative Results}
We evaluate our FLUX-Text on AnyText-benchmark~\cite{anytext2023} and MARIO-Eval-edit.
% Furthermore, we compare our approach with existing SoTA methods, including FLUX-Fill~\cite{flux2024}, AnyText~\cite{anytext2023}, and AnyText2~\cite{anytext22024}.
Furthermore, we compare our approach with existing SoTA methods, including DiffSTE~\cite{diff-ste2023}, TextDiffuser~\cite{textdiffuser2023}, DiffUTE~\cite{chen2023diffute}, AnyText~\cite{anytext2023}, TextCtrl~\cite{zeng2024textctrl}, and AnyText2~\cite{anytext22024}.

For AnyText-benchmark, as shown in Tab.~\ref{tab:quantitative_comparison}, FLUX-Text achieves the best performance across all metrics (Sen.ACC, NED, FID, LPIPS) on both Chinese and English datasets, despite being trained on only 100K samples compared to 2.9M samples used by AnyText.
For text fidelity, FLUX-Text reaches 84.19\% English Sen.ACC and 71.32\% Chinese Sen.ACC, surpassing AnyText2 by \textbf{+5.04\%} and \textbf{+1.10\%}, respectively. 
Notably, FLUX-Text delivers significant gains despite using only a small subset of the full training data and the inherent difficulty of Chinese text generation (large vocabulary and complex visual structures).
For generation quality, FLUX-Text attains the lowest FID (13.85 in English / 13.68 in Chinese) and LPIPS scores, indicating stronger distribution-level and perceptual alignment with ground-truth images. 
Qualitative results show that FLUX-Text maintains spatial consistency in edited regions while preserving unedited content, establishing robust multilingual performance, style consistency, and data efficiency that set a new state-of-the-art for scene text editing.

% \Cref{tab:mario} presents the results on the MARIO-Eval-edit benchmark, further confirming FLUX-Text’s advantage over other state-of-the-art methods.
\Cref{tab:mario} presents the results on the MARIO-Eval-edit benchmark. 
Our FLUX-Text achieves the best performance on the majority of metrics, including FID and all OCR-based measures, clearly outperforming AnyText and AnyText2.
When trained on the full dataset (marked as \( \ddagger \)), FLUX-Text further improves all metrics, achieving lower FID and higher CLIPScore, which demonstrates its strong generalization ability and data efficiency.
These results further confirm FLUX-Text’s advantage over other state-of-the-art methods.

\noindent \textbf{Qualitative Results}
As shown in~\Cref{fig:result_vis}, FLUX-Text produces more accurate and coherent text in both English and Chinese scenarios. It avoids the merged, missing, or distorted glyphs often seen in AnyText and AnyText2 and achieves better background consistency, while Flux-Fill fails to generate valid characters at all. 
We further validate the model's generalization on web-crawled images, as illustrated in~\Cref{fig:vis_zero} and~\Cref{fig:demo1}. 
These results highlight the strong overall performance and generalization ability of FLUX-Text, with strong text editing performance on these images.

\begin{table}[!t]
\centering
\caption{
% Encoder and Visual Embedding Type. Convs is for Multi-Encoders.
Visual Embedding Module is defined by the chosen Encoder and Visual Embedding type; Convs in “Encoder” correspond to Multi-Encoders.
}
\small
\setlength{\tabcolsep}{0.6mm}
\begin{tabular}{c|c|cc|cccc}
\toprule
\multirow{2}{*}{Idx} & \multirow{2}{*}{Encoder} & \multicolumn{2}{c|}{Type} & \multicolumn{2}{c}{English} & \multicolumn{2}{c}{Chinese} \\
                       &                                                                         & Canny             & Glyph             & Sen. ACC↑       & NED↑       & Sen. ACC↑       & NED↑      \\
\midrule
% (a)                    & \begin{tabular}[c]{@{}c@{}}Convs\\ (unfrozen)\end{tabular}              & $\times$          & \checkmark   &        &       &       &           \\
(a)                    & -                                                                       & $\times$          & $\times$     & 0.309  & 0.469 & 0.029 & 0.062   \\
\midrule
(b)                    &Convs                                                                    & $\times$          & \checkmark   & 0.613  & 0.775 & 0.067 & 0.135     \\
\midrule
(c)                    & \multirow{2}{*}{\begin{tabular}[c]{@{}c@{}}VAE\\ (frozen)\end{tabular}} & \checkmark        & $\times$     & 0.815  & 0.925 & 0.540 & 0.770     \\
(d)                    &                                                                         & $\times$          & \checkmark   & 0.800  & 0.922 & 0.617 & 0.805     \\
\bottomrule
\end{tabular}
\label{tab:ablation_visembed}
\end{table}

\begin{table}[!t]
\centering
\caption{
Ablation on the Text Embedding Module. 
% Note that the T5 encoder is included in all settings and thus omitted.
Note that the T5 encoder is included in all settings and is therefore omitted from the table for clarity.
}
\small
\setlength{\tabcolsep}{0.6mm}
\begin{tabular}{cc|cccc}
\toprule
\multicolumn{2}{c|}{Text Embedding} & \multicolumn{2}{c}{English} & \multicolumn{2}{c}{Chinese} \\
OCR\space\space\space\space          & Glyph-ByT5\space\space\space\space              & Sen. ACC↑       & NED↑       & Sen. ACC↑       & NED↑      \\
\midrule
% (a)                    & \begin{tabular}[c]{@{}c@{}}Convs\\ (unfrozen)\end{tabular}              & $\times$          & \checkmark   &        &       &       &           \\
$\times$          & $\times$     & 0.800  & 0.922 & 0.617 & 0.805   \\
\checkmark        & $\times$     & 0.779  & 0.910 & 0.591 & 0.786     \\
$\times$          & \checkmark   & 0.798  & 0.922 & 0.618 & 0.808     \\
\bottomrule
\end{tabular}
\label{tab:ablation_textembed}
\end{table}

\subsection{Ablation Study}
\label{sec:ablation_study}
% We build a small dataset of 100k images extracted from AnyWord-3M to perform the ablation. The dataset comprises 50k Chinese images and 50k English images. 
% Following the approach of previous methods, 
All ablation studies are systematically conducted on AnyText-benchmark~\cite{anytext2023}, with each setting trained for $15$K iterations for fairness.

\noindent \textbf{Visual Embedding.} 
As shown in~\Cref{tab:ablation_visembed}, comparing (a), (b), (c), and (d) underscores the significance of visual embedding in enhancing scene text editing.
The comparison between (b) and (d) shows that the choice of encoder has a clear impact on performance. 
We observe that the frozen VAE outperforms the trainable Multi-Encoder. 
With the VAE frozen, the comparison between (c) and (d) shows that, compared to glyph embedding, Canny embedding imposes overly strong spatial constraints on the injection of text-related visual information, reducing the effectiveness of Chinese text generation.
This is mainly because Chinese characters have highly diverse writing styles and are harder to spatially align with text-related visual information.
Overall, these results support our choice of using the frozen VAE with glyph embedding to form the Visual Embedding Module.

\noindent \textbf{Text Embedding.} 
\Cref{tab:ablation_textembed} compares different text embedding modules. 
Incorporating OCR learning embeddings led to a slight drop in performance, likely because the OCR-derived features primarily emphasize character-level recognition accuracy rather than holistic visual-semantic alignment. 
Such embeddings can be noisy when the OCR predictions are uncertain or erroneous, which may propagate errors into the DiT and degrade overall performance. 
Although Glyph-ByT5~\cite{glyphbyt5_2024} provides fine-grained semantic representations, it did not show significant gains in our experiments. 
This may be because Glyph-ByT5 relies on detailed text attributes such as font, language, and color, which were not sufficiently available in the actual test cases, limiting its advantages. 
By contrast, using only the T5 encoder achieves comparable performance and offers superior efficiency.

\noindent \textbf{Perceptual Loss.}  
As shown in~\Cref{tab:ablation_loss}, both the Text Perceptual (TP) loss, which serves as a global text perceptual loss, and the Regional Text Perceptual (RTP) loss lead to improved performance.
However, in scene text editing tasks, global perceptual loss often becomes less effective because large portions of the image remain unchanged. 
By focusing on edited text regions, our RTP loss consistently improves performance across metrics, including a \textbf{+2.4\%} gain in Chinese Sen. ACC, clearly demonstrating its advantage and effectiveness.

\noindent \textbf{Two-stage Training Strategy.}
To evaluate the effect of loss weight $\lambda$ in stage 2, we conduct an ablation study as shown in~\Cref{tab:loss_weight}, training each model for an additional $5$K iterations. 
When $\lambda=1$, the longer training only marginally improves performance over the baseline (stage 1). 
Increasing $\lambda$ progressively improves all metrics, with the best results achieved at $\lambda=30$. 
This indicates that assigning greater weight to text regions during stage 2 leads to better text-background integration and overall generation quality.

% \begin{table}[!t]
% \centering
% \caption{Ablation on Perceptual Loss.}
% \small
% \setlength{\tabcolsep}{0.6mm}
% \begin{tabular}{cc|cccc}
% \toprule
% \multicolumn{2}{c|}{Perceptual Loss} & \multicolumn{2}{c}{English} & \multicolumn{2}{c}{Chinese} \\
% TP          & RTP              & Sen. ACC↑       & NED↑       & Sen. ACC↑       & NED↑      \\
% \midrule
% % (a)                    & \begin{tabular}[c]{@{}c@{}}Convs\\ (unfrozen)\end{tabular}              & $\times$          & \checkmark   &        &       &       &           \\
% $\times$          & $\times$     & 0.698  & 0.865 & 0.564 & 0.767   \\
% \checkmark        & $\times$     & 0.802  & 0.922 & 0.591 & 0.797     \\
% $\times$          & \checkmark   & 0.798  & 0.922 & 0.618 & 0.808     \\
% \bottomrule
% \end{tabular}
% \label{tab:ablation_loss}
% \end{table}
\begin{table}[!t]
\centering
% \caption{Ablation on Perceptual Loss.}
\caption{Ablation on Perceptual Loss (TP: Textual Perceptual Loss, RTP: Regional Text Perceptual Loss).}
\small
\setlength{\tabcolsep}{0.6mm}
\begin{tabular}{cc|cccc}
\toprule
\multicolumn{2}{c|}{Perceptual Loss} & \multicolumn{2}{c}{English} & \multicolumn{2}{c}{Chinese} \\
TP          & RTP              & Sen. ACC↑       & NED↑       & Sen. ACC↑       & NED↑      \\
\midrule
% (a)                    & \begin{tabular}[c]{@{}c@{}}Convs\\ (unfrozen)\end{tabular}              & $\times$          & \checkmark   &        &       &       &           \\
$\times$          & $\times$     & 0.698  & 0.868 & 0.558 & 0.770   \\
\checkmark        & $\times$     & 0.798  & 0.921 & 0.593 & 0.793     \\
$\times$          & \checkmark   & \textbf{0.800}  & \textbf{0.922} & \textbf{0.617} & \textbf{0.805}     \\
\bottomrule
\end{tabular}
\label{tab:ablation_loss}
\end{table}

% \begin{table}[!t]
%     \centering
%     \small
%     \setlength{\tabcolsep}{1.2mm}
%     \caption{Ablation of loss weight $\lambda$ during stage 2 in a two-stage training strategy. The baseline denotes the stage 1 result. Bold numbers indicate the overall optimal $\lambda$.} 
%     \begin{tabular}{lcccc}
%         \toprule
%         \multirow{2}{*}{\text{Weight} $\lambda$} & \multicolumn{2}{c}{\text{Chinese}} & \multicolumn{2}{c}{\text{English}} \\
%         \cmidrule(lr){2-3} \cmidrule(lr){4-5}
%         & Sen.ACC\(\uparrow\) & NED\(\uparrow\) & Sen.ACC\(\uparrow\) & NED\(\uparrow\) \\
%         \midrule
%         Baseline  & 0.6171 & 0.8051 & 0.7996 & 0.9218 \\ \midrule
%         $\lambda=1$  & 0.6224 & 0.8114 & 0.8063 & 0.9225 \\
%         $\lambda=10$ & 0.6936 & 0.8392 & 0.8338 & 0.9342 \\
%         $\lambda=20$ & 0.7127 & 0.8513 & 0.8405 & 0.9386 \\
%         $\lambda=30$ & \textbf{0.7132} & \textbf{0.8510} & \textbf{0.8419} & \textbf{0.9400} \\
%         \bottomrule
%     \end{tabular}

%     \label{tab:loss_weight}
% \end{table}
\begin{table}[!t]
    \centering
    \caption{Ablation of loss weight $\lambda$ during stage 2 in a two-stage training strategy. The baseline denotes the stage 1 result. Bold numbers indicate the overall optimal $\lambda$.} 
    \small
    \setlength{\tabcolsep}{1.2mm}
    \begin{tabular}{lcccc}
        \toprule
        \multirow{2}{*}{\text{Weight} $\lambda$} & \multicolumn{2}{c}{\text{English}} & \multicolumn{2}{c}{\text{Chinese}} \\
        \cmidrule(lr){2-3} \cmidrule(lr){4-5}
        & Sen.ACC\(\uparrow\) & NED\(\uparrow\) & Sen.ACC\(\uparrow\) & NED\(\uparrow\) \\
        \midrule
        Baseline  & 0.7996 & 0.9218 & 0.6171 & 0.8051\\ \midrule
        $\lambda=1$  & 0.8063 & 0.9225 & 0.6224 & 0.8114 \\
        $\lambda=10$ & 0.8338 & 0.9342 & 0.6936 & 0.8392 \\
        $\lambda=20$ & 0.8405 & 0.9386 & 0.7127 & 0.8513 \\
        $\lambda=30$ & \textbf{0.8419} & \textbf{0.9400} & \textbf{0.7132} & \textbf{0.8510} \\  
        \bottomrule
    \end{tabular}
    \label{tab:loss_weight}
\end{table}

%% TODO tabel 引用有问题
%% table 和 图片 引用要统一（全称和简写要统一， 如 Fig., Figures, Tab., Table）
%% 主实验最好再补充
%% 文字问题：最好不要在同一段话里出现同一个单词

% \begin{table*}[ht]
%   \centering
%   \caption{Quantitative comparison of FLUX-Text and competing methods. 
%   \( \ddagger \) represent the model trained with full training dataset.}
%   \begin{tabular}{|l|c|c|c|c||c|c|c|c|}
%   \hline
        % \toprule
        % Exp. & Embedding Granularity & \begin{tabular}[c]{@{}c@{}}Embedding Modal \\ Visual\end{tabular} & \begin{tabular}[c]{@{}c@{}}Embedding Modal \\ Text\end{tabular} & \begin{tabular}[c]{@{}c@{}}Perceptual Loss \\ OCR\end{tabular} & \begin{tabular}[c]{@{}c@{}}Perceptual Loss \\ ODM\end{tabular} & $\lambda$ & \begin{tabular}[c]{@{}c@{}}Sen. \\ ACC $\uparrow$\end{tabular} & \begin{tabular}[c]{@{}c@{}}NED \\ $\uparrow$\end{tabular} \\ \midrule

\section{Conclusion}
\label{sec:conclusion}
In this paper, we present FLUX-Text, a novel DiT-based method for multilingual scene text editing. 
By integrating the Visual and Text Embedding Modules into the DiT architecture, FLUX-Text effectively injects text-editing cues, which is critical for text editing.
The proposed Regional Text Perceptual loss and two-stage training strategy further enhance text fidelity by focusing the model's attention on text regions, ensuring harmonious integration with diverse backgrounds.
Compared to existing methods, FLUX-Text achieves state-of-the-art performance with only $100$K training examples (a \textbf{97\%} reduction in data requirements), demonstrating remarkable efficiency and scalability across both Latin and non-Latin text editing tasks. Notably, FLUX-Text pioneers the integration of DiT architectures into scene text editing, setting a new benchmark for quality and adaptability, and we hope it inspires further exploration of DiT-based approaches in this domain.

{
    \small
    \bibliographystyle{ieeenat_fullname}
    \bibliography{main}
}

\end{document}